\documentclass[lettersize,journal]{IEEEtran}
\usepackage{amsmath,amsfonts}
\usepackage{algorithmic}
\usepackage{array}
\usepackage[caption=false,font=normalsize,labelfont=sf,textfont=sf]{subfig}
\usepackage{textcomp}
\usepackage{stfloats}
\usepackage{url}
\usepackage{verbatim}
\usepackage{graphicx}
\usepackage{booktabs}
\usepackage{graphicx}
\usepackage{pifont}
\usepackage{changepage}
\usepackage{pifont}
\usepackage{xcolor}
\usepackage{multirow}
\usepackage{multicol}
\usepackage{caption}
\usepackage[linesnumbered,ruled,vlined]{algorithm2e}
\def\BibTeX{{\rm B\kern-.05em{\sc i\kern-.025em b}\kern-.08em
    T\kern-.1667em\lower.7ex\hbox{E}\kern-.125emX}}
\usepackage{balance}
\begin{document}
\title{AICL: Action In-Context Learning for \\
Video Diffusion Model}
\author{
    Jianzhi Liu,
    Junchen Zhu,
    Lianli Gao,
    Heng Tao Shen, IEEE Fellow,
    and Jingkuan Song
}

\markboth{IEEE TRANSACTIONS ON IMAGE PROCESSING,~Vol. ~x, N0. ~x, July~2024}
{How to Use the IEEEtran \LaTeX \ Templates}



\twocolumn[{%
\renewcommand\twocolumn[1][]{#1}%
\maketitle
\begin{center}
    \centering
    \includegraphics[width=\textwidth]{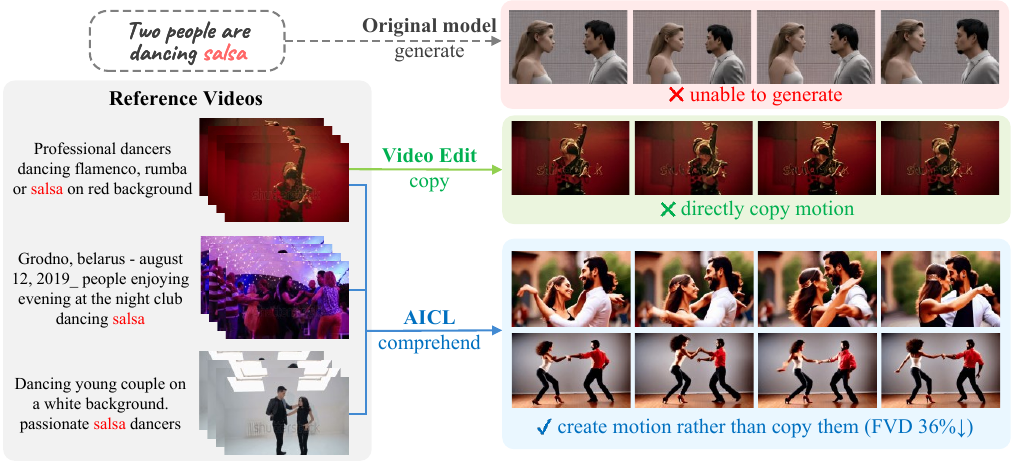}
    \captionof{figure}{Some video generation models struggle to generate certain actions. For example, the \textbf{original model} cannot understand and generate ``salsa''. The \textbf{video edit} methods only directly copy motion from reference videos. The model with \textbf{AICL} (Action In-Context Learning) can \textit{comprehend} relevant videos without training on this specific action with a few videos as references.}
    \label{Fig: banner}
\end{center}%
}]


\begin{abstract}



The open-domain video generation models are constrained by the scale of the training video datasets, and some less common actions still cannot be generated. Some researchers explore video editing methods and achieve action generation by editing the spatial information of the same action video. However, this method mechanically generates identical actions without understanding, which does not align with the characteristics of open-domain scenarios. In this paper, we propose AICL, which empowers the generative model with the ability to understand action information in reference videos, similar to how humans do, through in-context learning. Extensive experiments demonstrate that AICL effectively captures the action and achieves state-of-the-art generation performance across three typical video diffusion models on five metrics when using randomly selected categories from non-training datasets.
\end{abstract}


\section{Introduction}

\begin{adjustwidth}{0.7cm}{0.7cm}
    \textit{``Imitation is the sincerest form of flattery that mediocrity can pay to greatness.''\ \ \ --- Oscar Wilde~\cite{wilde1894}}
\end{adjustwidth}

The advent of large-scale generative models has significantly propelled the field of video generation forward in recent years~\cite{LVDM, chen2023videocrafter1, chen2024videocrafter2, wang2023modelscope, ohya2013automatic, yang2023enabling, satapathy2023video}. Among various architectures, the diffusion model~\cite{Diffusion} has emerged as the predominant choice due to its proficiency in generation quality and stability during training. Video Diffusion Models (VDMs) integrate two critical components: Spatial layers that generate and process the shapes within frames, and Temporal layers that maintain consistency across frames. Given that images, which lack temporal dimensions, are relatively simpler for models to master, many video diffusion models extend from pre-trained image diffusion models~\cite{HongDZLT23CogVideo, HoCSWGGKPNFS22ImagenVideo, WuGWLGHSQS22tuneavideo}. This extension leverages the exceptional spatial generation capabilities of image models~\cite{DingZHT22cogview2, imagen, Stable-Diffusion}. However, adapting these models to capture complex temporal dynamics remains a challenge, with some approaches focusing on generating high-quality outputs by modeling a single action~\cite{wu2023tune}. For more practical applications, such as open-domain video generation, most works suffer from the limited model scale in various temporal modelings as in Fig.\ref{Fig: banner}.

The limitations in scaling models for enhanced temporal ``learning'' have led researchers to explore ``imitation'' strategies, focusing on extracting and learning from pertinent information in examples, rather than solely relying on training data. This approach, inspired by the success of in-context learning in natural language processing (NLP) tasks~\cite{wei2023symbol, gu2023pre, li2023finding, xu2023small, xu2023k}, leverages Large Language Models (LLMs) to make predictions based on a few examples. Current efforts in multi-modal in-context learning primarily address simpler image or audio modalities, either by developing models that extract and align features across modalities or by integrating encoded visual modalities with tokens through auto-encoders~\cite{ding2022gpt, huang2023language, tsimpoukelli2021multimodal, sun2023exploring, bar2022visual, wang2023images, moens2021findings, zhang2023makes}. Despite the promising results in text, image, and audio, extending in-context learning to the complex video domain poses significant challenges. \cite{hertz2022prompt} investigates the influence of spatial attention in generating spatial contents, showing that altering the attention map for specific prompt words can change the spatial features of the generation. Using this method can optimize the spatial information in generated videos. However, this method has not addressed the essential part of videos, i.e., temporal content or action.

To address these challenges, we introduce a novel framework in this paper, namely \textit{AICL}, which leverages the abundant availability of video data to enhance action generation in pre-trained and frozen VDMs through an in-context learning approach. AICL incorporates an Action Prism (AP) that aligns input conditions (e.g., text descriptions) with video content to effectively extract relevant motion features. AICL then injects these motion features into the pipeline of pretrained VDMs, which is achieved by introducing new attention layers that integrate these motion features through cross-attention mechanisms. AICL is trained with a small amount of data and only uses a single reference video for training but can flexibly incorporate multiple references during inference to produce the desired output, showcasing its versatility. Remarkably, AICL achieves superior performance in generating diverse actions, including those not seen during training, without directly replicating visual content from reference videos. This capability is validated through extensive experiments, highlighting our significant advancements over existing methods for diverse high-quality action generation without fine-tuning.

Our contributions in this paper can be summarized as follows:
\begin{itemize}
\item We introduce AICL, an innovative method designed to augment the capabilities of existing video diffusion models, enabling them to synthesize complex actions beyond their original scope using in-context learning. It can utilize multiple reference videos to achieve a broader spectrum of action imitation and generate novel actions without fine-tuning.
\item We introduce Action Prism to capture visual motion features for directing the generation of video diffusion models. It can distill effective and related visual motion features instead of replicating the referred content.
\item We perform experiments on 3 typical baselines and show that AICL enhances the action generation capabilities of current video diffusion models and significantly helps to decrease the FVD by 36\%. A generalization study proves that AICL can be easily transferred to image-to-video tasks.
\end{itemize}

\section{Related Works}
\subsection{Video Diffusion Model}

Diffusion models~\cite{Diffusion} have demonstrated significant advancements in the domain of visual generation. To address the computational challenges associated with high-resolution pixel space, researchers have shifted towards latent space diffusion processes, culminating in the development of the Latent Diffusion Model (LDM)~\cite{Stable-Diffusion}. Leveraging insights gained from latent image diffusion models~\cite{Stable-Diffusion,imagen,2023dalle3}, recent efforts have expanded into the realm of latent video diffusion models. Building upon successful text-based image diffusion models, analogous text-based video diffusion strategies have emerged. Notably, MagicVideo~\cite{MagicVideo} pioneered the use of LDM for video generation within the latent space, achieving a significant reduction in computational demands. It introduced a novel frame-wise Temporal Attention mechanism to ensure consistency across video frames. Subsequently, LVDM~\cite{LVDM} also utilized LDM for spatial representation and incorporated Conditional Latent Perturbation~\cite{Conditional-Latent-Perturbation} and Unconditional Guidance~\cite{Diffusion-Guidance} to mitigate performance declines.
VideoFactory~\cite{VideoFactory} introduced an innovative attention mechanism that alternates the roles of spatial and temporal features within the attention module, improving spatial-temporal modeling.

Expanding the utility of video diffusion models, research has progressively explored multi-modal guided video generation. VideoComposer~\cite{wang2023videocomposer} introduced a Spatio-Temporal encoder capable of flexibly extracting multi-modal features, thereby enhancing the quality of generated video content. 
MovieFactory~\cite{zhu2023moviefactory} integrated ChatGPT~\cite{openai2023gpt4} for script generation and an audio retrieval framework for voice-overs, facilitating the creation of comprehensive audio-visual content. However, these methods typically require training on specific action categories, leading to significant performance degradation when encountering previously unseen actions. 

\subsection{In-Context Learning (ICL)}

The emergence of Large Language Models (LLMs) has significantly advanced machine learning, particularly showcasing the potential of In-Context Learning (ICL) to facilitate task mastery with minimal examples. 
Researchers have explored various in-context fine-tuning strategies, such as creating tailored in-context learning datasets~\cite{wei2021finetuned, wang2022super, isabelle2002proceedings, gu2023pre} and introducing MetaICL to seamlessly transition from pretraining to application~\cite{chen2022improving, min2021metaicl}. 
Inference effectiveness largely hinges on the judicious selection of demonstration examples. Techniques like the kNN-based unsupervised retriever method have been developed to enhance example selection, thus optimizing the model's learning from these instances~\cite{rubin2021learning, li2023finding, magister2022teaching}. 

Building on NLP successes, ICL's applicability is being extended to various modalities, especially in vision-language integration. Developments like image embedding sequences and Flamingo, a multi-modal LLM plug-in, underscore the efficacy of ICL in few-shot learning scenarios across different tasks and modalities, leveraging large-scale multi-modal web data~\cite{tsimpoukelli2021multimodal, alayrac2022flamingo}. 
Nevertheless, the intricacy of video temporal features makes ICL for video action a challenging endeavor.
This paper specifically focuses on employing ICL for video generation, aiming to enhance the creation of complex actions by imitating existing videos, thereby addressing gaps in pretrained models' capabilities.

\begin{figure*}[t]
	\centering
	\includegraphics[width=\textwidth]{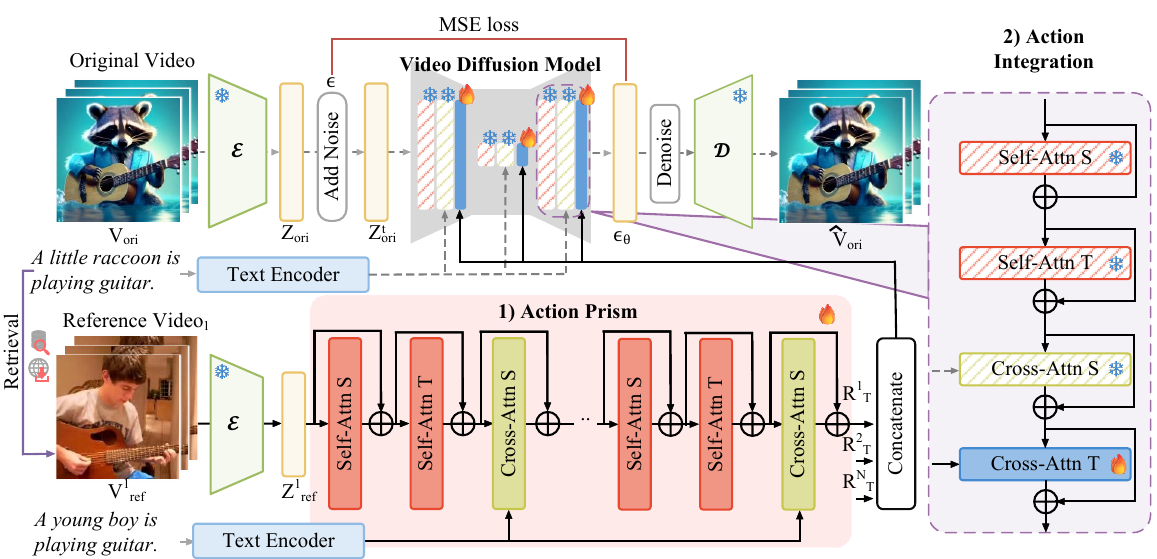}
    \caption{Illustration of the proposed AICL for text-to-video diffusion models. AICL contains two parts: 1) \textbf{Action Prism}, marked with \textcolor{magenta}{pink shading}, to extract action features, and 2) \textbf{Action Integration}, marked with \textcolor{violet}{purple shading}, to integrate action features. S and T stand for spatial and temporal respectively.}

	\label{Fig: overview}
\end{figure*}

\section{Method}

In this section, we begin by introducing some preliminary concepts in AICL and formulate essential models and parameters in Sec.~\ref{Sec: Preliminary}. Subsequently, we present our Action Prism and Action Integration in Sec.~\ref{Sec: Action Prism} and Sec.~\ref{Sec: Inject Method}. Building upon the model, we delineate the selection of reference videos, training, and inference in Sec.~\ref{Sec: Pipeline}.

\subsection{Preliminary} \label{Sec: Preliminary}
Diffusion model~\cite{Diffusion} is a generation model containing two processes: 1) forward process to add noise to the clean sample with T steps to make it a Gaussian noise, and 2) backward process to predict and remove the noise of the current sample which is also repeated for T times thus obtaining the clean generated sample.
Latent Video Diffusion Models, which follow the basic diffusion model, consist of two main components: Variational Autoencoder (VAE) and Diffusion U-Net. The VAE compresses the original video $V_{ori}$ into latent space and can convert the latent feature back to pixel spaces as follows:
\begin{equation}
    z_{ori}=\varepsilon (V_{ori}),\quad \widehat{V}_{ori}= \mathcal{D}(z_{ori}),
\end{equation}
where $\varepsilon$ and $\mathcal{D}$ indicate the encoder and decoder of the VAE, respectively. The diffusion U-Net network is used to predict the noise $\epsilon_\theta (z_{{ori}_t}, C_{ori}, t)$ of the current input latent sample in the timestep $t$ based on the paired condition of the video $C_{ori}$ (usually encoded, e.g., the text condition is usually encoded by the CLIP text encoder~\cite{Clip}), where $\theta$ indicates the parameter of the network.

Diffusion U-Net networks work with two attention mechanisms~\cite{VaswaniSPUJGKP17}: self-attention (SA) to improve the representation of the feature and cross-attention (CA) to inject information from other features (e.g., text). The queries $Q$, keys $K$, and values $V$ for self-attention are derived from the same source, while for cross-attention, $Q$ and $K, V$ originate from different sources. Each feature for calculation will be transformed by an independent matrix $W$. The specific formulations for $SA$ and $CA$ can be described as follows:
\begin{equation}
\begin{aligned}
    SA(x) &= \text{softmax}\left(\frac{W_Q(x)W_K(x)^T}{\sqrt{d_{W_K(x)}}}\right)W_V(x),  \\
    CA(x, y) &= \text{softmax}\left(\frac{W_Q(x)W_K(y)^T}{\sqrt{d_{W_K(y)}}}\right)W_V(y),    
\end{aligned}
\end{equation}

In this paper, we introduce extra reference videos $V^i_{ref}$ with their corresponding condition information $C^i_{ref}$ to direct the diffusion process of the original video. Following the latent video diffusion model to enhance model efficiency and minimize parameter count, we use the latent feature of $V^i_{ref}$ by encoding them with the pretrained $\varepsilon$ as follows:
\begin{equation}
    z^i_{ref}=\varepsilon (V^i_{ref}).
\end{equation}

\subsection{Action Prism} \label{Sec: Action Prism}


Action Prism (AP) is introduced to extract action-related visual information from reference videos. Drawing inspiration from the pioneering work~\cite{VaswaniSPUJGKP17}, our model is constructed on the foundation of a transformer-based architecture. As depicted in Fig.\ref{Fig: overview}, each block of the AP incorporates three distinct attention layers to process the data. Unlike traditional transformers that handle 1-dimensional textual data, videos represent a more complex 3-dimensional data structure, encompassing both spatial dimensions (height $h$ and width $w$) and a temporal dimension (frame length $D$). The naive approach of flattening all channels would not only be computationally prohibitive but also pose significant optimization challenges. To address this, we decouple the processing of spatial and temporal information through two separate self-attention layers.

For the spatial self-attention layer, we reorganize the input features by considering the video frame length as part of the batch dimension, while flattening the spatial dimensions (height and width) into the sequence length dimension. Conversely, in the temporal self-attention mechanism, the spatial dimensions are grouped into the batch dimension, and the frame length is treated as the sequence length. To enhance the extraction process with additional contextual information, we incorporate a cross-attention mechanism. This mechanism is designed with a similar separation principle, distinguishing between spatial and temporal aspects. Interestingly, our experimental findings suggest that temporal cross-attention does not contribute to motion extraction in such a design. Consequently, we exclusively employ spatial cross-attention to combine the condition information. 
Each block of our AP can be formulated as follows:
\begin{equation}
    \begin{aligned}
        h_1 &= SA_{T}(SA_{S}(h_0)),  \\
        h_2 &= CA_{S}(h_1, C^i_{ref}), \\
        h_3 &= \text{MLP}(h_2),
    \end{aligned}
\end{equation}
where $h_0$ is the input of this block, and $h_3$ is the output. MLP is a two-layer fully connected network. Then, we can describe the function of the whole MP as follows:
\begin{equation}
    R^i=\text{AP}(z^i_{ref},C^i_{ref}).
\end{equation}
To integrate the information from several reference videos we simply concatenate all $R^i$ which we find works well in our experiment as follows:
\begin{equation}
    R = \prod_{i=1}^{k}R^i,
\end{equation}
where $\prod$ indicates the concatenation operation and k represents the number of the reference videos. This structured approach enables the AP to efficiently and effectively distill action-specific features from complex video data. Then we can use these features to direct the action generation of original VDMs.

\subsection{Action Integration} \label{Sec: Inject Method}
In this study, our objective is to facilitate the generation of diverse actions without necessitating specific, subsequent fine-tuning. To achieve this, the method of integration must not compromise the integrity of the original layers within the pre-trained VDMs, as tampering with these layers could lead to a loss of the extensive knowledge they encapsulate. In light of this, we advocate for the introduction of new layers to the diffusion network, while maintaining all pre-existing, pre-trained layers in a frozen state. This approach ensures that the foundational knowledge within the VDM remains intact and unaltered. Our experimental analyses reveal an interesting observation: spatial cross-attention does not significantly enhance visual generation, instead it can reduce the coherence of action generation. This insight has led us to solely utilize temporal cross-attention for guiding the temporal motion. As illustrated in Fig.\ref{Fig: overview}, we incorporate a temporal cross-attention layer at the end of each UNet block. The function of a UNet block, with our modifications, can be represented by the following equations:
\begin{equation}
    \begin{aligned}
        h_1 &= SA_{T}(SA_{S}(h_0)), \\
        h_2 &= CA_{T}(CA_{S}(h_1), C_{ori}), \\
        h_3 &= CA'_{T}(h_2, R),
    \end{aligned}
    \label{Formula: inject}
\end{equation}
where $CA'_T$ denotes the newly added temporal cross-attention layer. Note that all other layers within this framework are fixed during the training. This strategic addition and preservation approach allows us to harness the pre-trained VDMs for action generation while ensuring the new layers adapt to inject the motion information from reference videos without compromising the pre-existing, valuable knowledge embedded within the network.


\begin{algorithm}[t!]
\SetKwInOut{Input}{input}
\SetKwInOut{Output}{output}
\SetKwFunction{fun}{AP}
\SetKwProg{Fn}{Function}{:}{}

\Input{$C_{ori}, V_{ori}$, $(C^1_{ref}, V^1_{ref})$}
\Output{$loss$}
\BlankLine 
$z^1_{ref} = \varepsilon(V^1_{ref})$\;
$R = $AP$(z^1_{ref}, C^1_{ref})$\;
$t \leftarrow U(1,T)$\;
$\epsilon \leftarrow \mathcal{N}(0,1)$\; 
$z_{ori} = \varepsilon(V_{ori})$ \;
$z_{ori_t} = \sqrt{\bar{\alpha}_t}z_{ori} + \sqrt{1 - \bar{\alpha}_t}\boldsymbol{\epsilon}$\;
$loss = \mathcal{L}_{simple}(\epsilon, \epsilon_\theta (z_{{ori}_t}, C_{ori}, t, R))$

\caption{The training pipeline of AICL. We only take a single reference video for each sample when training. $\bar{\alpha}_t$ is the hyper-parameter of the diffusion process~\cite{Diffusion}.}
\label{Algorithm: Pipeline-training}
\end{algorithm}

\begin{algorithm}[t!]
\SetKwInOut{Input}{input}
\SetKwInOut{Output}{output}
\SetKwFunction{fun}{AP}
\SetKwProg{Fn}{Function}{:}{}

\Input{$C_{ori}$, $\{(C^1_{ref}, V^1_{ref}), \cdots,(C^k_{ref}, V^k_{ref})\}$}
\Output{$V_{out}$}
\BlankLine 
\For{$i \leftarrow 1$ \KwTo $k$}{
    $z^i_{ref} = \varepsilon(V^i_{ref})$\;
    $R^i = $AP$(z^i_{ref}, C^i_{ref})$\;
}
$R = \prod_{i=1}^{k}R^i$\;
$z_T \leftarrow \mathcal{N}(0,1)$ \;
\For{$t \leftarrow T$ \KwTo $1$}{
    $z_{t - 1} = Diffusion(z_{t}, C_{ori}, t, R)$\;
}
$V_{out} = \mathcal{D}(z_{0})$ \;

\caption{The inference pipeline of AICL. AICL supports multiple reference videos for inference. $Diffusion$ indicates the existing diffusion algorithms, like DDPM~\cite{ho2020ddpm} and DDIM~\cite{song2020ddim}.}
\label{Algorithm: Pipeline-inference}
\end{algorithm}





\subsection{Pipeline} \label{Sec: Pipeline}
With all modules described above, we delineate the comprehensive pipeline of our methodology, which encompasses the selection of reference videos, training, and inference processes.

\subsubsection{Selection of Reference Videos} The choice of reference videos is pivotal to the efficacy of our framework. The relevance of the reference videos to the user input significantly enhances the ability of our model to extract pertinent motion features. Our research primarily focuses on one prevalent type of video generation: text-to-video, which requires the model to synthesize complete motion sequences. Specifically, we consider two typical data. First, many datasets lack descriptive textual annotations and are often constrained to specific scenarios. We manually annotate these datasets with action descriptions. For instance, for a collection of videos depicting the action ``playing guitar'', we assign a uniform prompt such as "A person is playing guitar" to the entire cluster. Second, for datasets that are open-domain and include a wide range of videos with potentially noisy textual descriptions, we opt to utilize the original captions without any preprocessing. To select appropriate references, we extract the principal verbs or phrases, denoted as $V$, from the captions across all videos and construct a comprehensive list of verbs $L = \{V_1, V_2, \ldots, V_k\}$. This list includes various verb forms, encompassing both continuous and past tenses. Reference videos are then selected based on the following criteria:
\begin{equation}
\exists_v( v \in C_{ori} \land v \in C_{ref} \land v \in L),
\end{equation}
It's important to note that videos satisfying this condition are chosen randomly as references, without any additional filtering.

\subsubsection{Training and Inference} The training and inference algorithms are demonstrated in Algorithm.\ref{Algorithm: Pipeline-training} and Algorithm.\ref{Algorithm: Pipeline-inference}. The algorithms are similar to those of original diffusion, except that we introduce the reference videos in the process. In training, we follow the basic loss function to train our network as follows:
\begin{equation}
    \mathcal{L}{simple}(\epsilon, \epsilon_\theta) = \left \| \epsilon - \epsilon_\theta \right \|^2_2.
\end{equation}
It is noteworthy that our findings indicate that training the model with a single reference video is adequate for it to generalize effectively to scenarios involving multiple references, without necessitating any additional fine-tuning. During inference, the Action Prism only executes a singular forward pass, and the extracted feature is applied across the entirety of the diffusion timestep sequences. As a result, the integration of AICL into the diffusion process incurs a marginal increase in inference time, approximately 0.5\% when employing a 50-step diffusion algorithm with one reference video. This efficiency confirms the practicality of incorporating AICL into existing VDMs, offering enhanced motion feature extraction with minimal impact on computational costs.
\begin{figure*}[]
        \centering
	\includegraphics[]{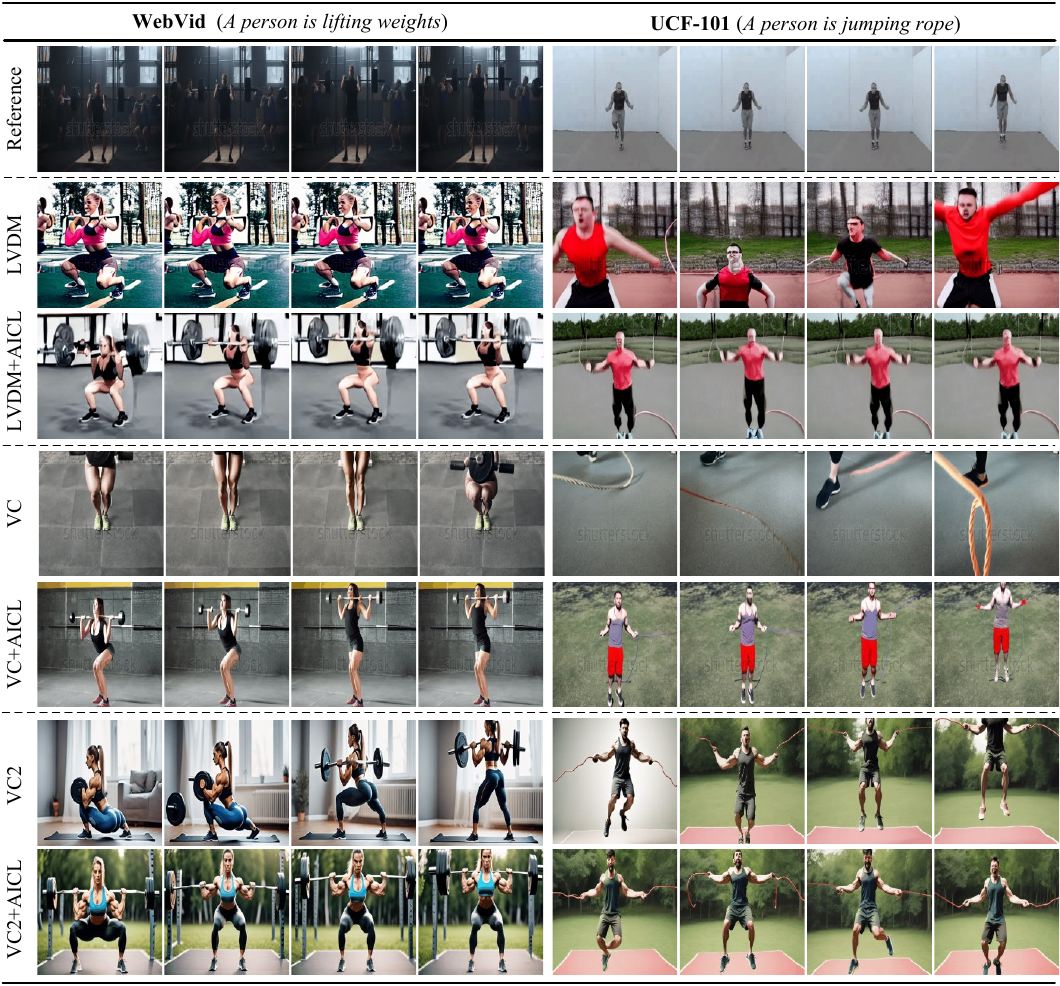}
	\caption{Qualitative results of comparison between baseline models and AICL with single reference video.}
	\label{Fig: Main-Result}
    \vspace{-1em}
\end{figure*}

\section{Experiments}
In this section, we first introduce our settings in Sec.~\ref{Sec: Settings}. Next, we validate the effectiveness and generalization ability of AICL in Sec.~\ref{Sec: Main Results} and Sec.~\ref{Sec: Generalization Study}. Finally, we present ablation studies to analyze the impacts of different factors on AICL in Sec.~\ref{Sec: Ablation Study}.

\subsection{Settings} \label{Sec: Settings}
\noindent\textbf{Baselines.} 
We choose three state-of-the-art open-source video diffusion models as our baselines, including LVDM~\cite{LVDM}, VideoCrafter (VC) ~\cite{chen2023videocrafter1}, and VideoCrafter2 (VC2) ~\cite{chen2024videocrafter2}. All models support the text-to-video and VideoCrafter provides the application for image-to-video.


\noindent\textbf{Datasets.} 
We utilize two datasets as zero-shot test settings to measure AICL: UCF-101~\cite{UCF} and WebVid~\cite{WebVid}. UCF-101 contains 101 actions, and we select 10 poorly generated actions for testing and the other 91 actions for training. WebVid is only used for testing without any fine-tuning. We randomly choose 1,000 videos from it and make sure their actions are not included in the training set of UCF-101.


\noindent\textbf{Evaluation metrics.} 
To comprehensively evaluate the effectiveness of AICL, we adopt five commonly used metrics. We utilize a) Clip~\cite{Clip} to measure the text-visual alignment and coherence of generated videos denoted as \textbf{CC} (Cross-model Consistent) and \textbf{FC} (Frames Consistent), b)  \textbf{FVD}~\cite{FVD} to evaluate the feature distribution between real and synthesized videos, c) User Study to evaluate the Cross-model Consistent and Frames Consistent as \textbf{USCC} and \textbf{USFC}. We carried out a randomized assessment involving 10 evaluators, all of whom have artistic backgrounds and are engaged in photography, video editing, painting, etc.

\noindent \textbf{Training Details.}
For hyperparameters, we set $\lambda_{\mathcal{L}_{vlb}}=0$ to disable vlb loss. Deep Speed ~\cite{rasley2020deepspeed} CPU Adam is adopted in all experiments, with learning rate $lr=1e^{-4}$, betas $\beta=(0.9, 0.9)$ and weight decay $0.03$ for all components. We use optimizer and gradient state partitioning for the Deep Speed strategy. We keep the batch size at $6$ for LVDM training, batch size at $4$ for VideoCrafter T2V training, and batch size at $3$ for VideoCrafter I2V training. For all experiments, we train a total of 160k text-video-reference or text-image-video-reference pairs. We train LVDM with 256 $\times$ 256 resolution videos, and train VideoCrafter with 512 $\times$ 320 resolution videos. Mixed precision training from native Deep Speed is adopted (has significant speed improvement in training and minor in testing). In addition, we use memory efficient cross attention of xformers ~\cite{xFormers2022} to boost the training and inference step. All experiments are conducted on NVIDIA RTX 3090 GPUs.

\noindent \textbf{Model Details.}
The Action Prism is designed to extract action-related visual information from reference videos. It utilizes an 18-layer transformer architecture for processing. The input data undergoes an initial transformation through a multi-layer perceptron (MLP) to adjust the channel dimensions. Specifically, the video channel dimension is expanded from 4 to 640, and the text channel dimension is decreased from 1024 to 768. The transformer architecture is then applied, with each layer comprising three types of attention: spatial self-attention, temporal self-attention, and spatial cross-attention. These attention are configured with 8 heads, a head dimension of 80, and they utilize relative positioning. Spatial self-attention employs the xformers for processing. Then the data is passed through a ``to out'' layer. This layer consists of 3 MLP layers, which initially expand the channel dimension by a factor of four, process the data at this expanded dimension, and then reduce the channel dimension back to 4. This processed output is then injected into a U-Net architecture for further processing.

\begin{table*}[]
    \caption{Quantitative results of comparison between baseline models and AICL with single reference video.}
    \centering{
        \begin{tabular}{@{}l|ccccc|ccccc@{}}
\toprule
\multirow{2}{*}{Method} & \multicolumn{5}{c|}{WebVid}                                                                                & \multicolumn{5}{c}{UCF-101}                                                                                \\
                        & CC $\uparrow$  & FC $\uparrow$  & FVD $\downarrow$ & USCC $\uparrow$ & USFC $\uparrow$ & CC $\uparrow$  & FC $\uparrow$  & FVD $\downarrow$ & USCC $\uparrow$ & USFC $\uparrow$ \\ \midrule
LVDM                    & \textbf{30.42} & 94.53          & 1349             & 18.75\%                                    & 6.25\%                                     & 27.22          & 91.88          & 1267             & 39.58\%                                    & 20.83\%                                    \\
LVDM + AICL                    & 30.41          & \textbf{97.46} & \textbf{998}     & \textbf{81.25\%}                           & \textbf{93.75\%}                           & \textbf{28.72} & \textbf{96.86} & \textbf{897}     & \textbf{60.42\%}                           & \textbf{79.17\%}                           \\ \midrule
VC            & 31.37          & 96.23          & 938              & 5.26\%                                    & 5.26\%                                    & 27.06          & 93.02          & 1316             & 25.00\%                                    & 9.09\%                                    \\
VC + AICL                    & \textbf{32.46} & \textbf{98.44} & \textbf{798}     & \textbf{94.74\%}                           & \textbf{94.74\%}                           & \textbf{27.94} & \textbf{95.28} & \textbf{1106}    & \textbf{75.00\%}                           & \textbf{90.91\%}                           \\ \midrule
VC2           & 31.18          & 95.03          & 1456             & 14.29\%                                    & 16.67\%                                    & 27.88          & 91.15          & 1122             & 9.09\%                           & 10.00\%                                    \\
VC2 + AICL                    & \textbf{33.28} & \textbf{96.57} & \textbf{1195}    & \textbf{85.71\%}                           & \textbf{83.33\%}                           & \textbf{28.13}    & \textbf{93.46} & \textbf{997}    & \textbf{90.91\%}                           & \textbf{90.00\%}                           \\ \bottomrule
\end{tabular}
    }
\label{Table: Main-Results}
\end{table*}

\begin{table*}[t]
\centering
\caption{Quantitative results of ablation study of AICL applied to LVDM using UCF dataset.}
\begin{tabular}{@{}c|cccccc|cc|ccc@{}}
\toprule
\multirow{3}{*}{Model} & \multicolumn{6}{c|}{Action Prism}                                                           & \multicolumn{2}{c|}{Action Integration} & \multicolumn{3}{c}{\multirow{2}{*}{Metrics}}     \\ \cmidrule(lr){2-9}
                       & \multicolumn{2}{c|}{Block}                  & \multicolumn{4}{c|}{Cross-Attention}              & \multicolumn{2}{c|}{Added Attention}   & \multicolumn{3}{c}{}                             \\ \cmidrule(l){2-12} 
                       & Self-Attn & \multicolumn{1}{c|}{Cross-Attn} & Q & \multicolumn{1}{c|}{K/V} & Spatial & Temporal & Spatial           & Temporal           & CC $\uparrow$ & FC $\uparrow$ & FVD $\downarrow$ \\ \midrule
LVDM                               & -            & \multicolumn{1}{c|}{-}            & -     & \multicolumn{1}{c|}{-}     & -            & -            & -                  & -                 & 27.22          & 91.88          & 1267             \\ \midrule
\textcircled{1} &              & \multicolumn{1}{c|}{$\checkmark$} & Video & \multicolumn{1}{c|}{Text}  & $\checkmark$ &              &                    & $\checkmark$      & 28.49          & 96.20          & 1084             \\
\textcircled{2} & $\checkmark$ & \multicolumn{1}{c|}{$\checkmark$} & Text  & \multicolumn{1}{c|}{Video} & $\checkmark$ &              &                    & $\checkmark$      & 24.90          & \textbf{97.19} & 1136             \\
\textcircled{3} & $\checkmark$ & \multicolumn{1}{c|}{$\checkmark$} & Video & \multicolumn{1}{c|}{Text}  & $\checkmark$ &              &                    & $\checkmark$      & \textbf{28.72} & 96.86          & \textbf{897}     \\ \midrule
\textcircled{4} & $\checkmark$ & \multicolumn{1}{c|}{$\checkmark$} & Video & \multicolumn{1}{c|}{Text}  & $\checkmark$ & $\checkmark$ &                    & $\checkmark$      & 27.98          & 95.21         & 1122             \\
\textcircled{5} & $\checkmark$ & \multicolumn{1}{c|}{$\checkmark$} & Video & \multicolumn{1}{c|}{Text}  &              & $\checkmark$ &                    & $\checkmark$      & 28.11          & 95.36         & 1068             \\
\textcircled{6} & $\checkmark$ & \multicolumn{1}{c|}{$\checkmark$} & Video & \multicolumn{1}{c|}{Text}  & $\checkmark$ &              &                    & $\checkmark$      & \textbf{28.72} & \textbf{96.86} & \textbf{897}     \\ \midrule
\textcircled{7} & $\checkmark$ & \multicolumn{1}{c|}{$\checkmark$} & Video & \multicolumn{1}{c|}{Text}  & $\checkmark$ &              & $\checkmark$       &                   & 28.48          & \textbf{97.54} & 1049             \\
\textcircled{8} & $\checkmark$ & \multicolumn{1}{c|}{$\checkmark$} & Video & \multicolumn{1}{c|}{Text}  & $\checkmark$ &              & $\checkmark$       & $\checkmark$      & 27.17          & 97.04          & 1052             \\
\textcircled{9} & $\checkmark$ & \multicolumn{1}{c|}{$\checkmark$} & Video & \multicolumn{1}{c|}{Text}  & $\checkmark$ &              &                    & $\checkmark$      & \textbf{28.72} & 96.86          & \textbf{897}     \\ \bottomrule
\end{tabular}
\label{Table: ablation}
\end{table*}

\subsection{Main Results} \label{Sec: Main Results}
\noindent \textbf{Compare With Other SOTA.} To evaluate the efficiency of AICL, we conduct extensive experiments on WebVid and UCF-101 datasets. 
As can be observed from Table \ref{Table: Main-Results}, AICL contributes to significant improvements across nearly all metrics. In terms of objective assessments, the integration of AICL into models results in substantial reductions FVD. For instance, a decline from 1349 to 998 for LVDM confirms AICL's capability to generate high-quality motion. From the perspective of subjective evaluation, AICL is recognized for its considerable improvements in ensuring text-visual alignment and enhancing frame consistency. AICL's performance on the open-domain WebVid dataset surpasses its efficacy on the smaller UCF-101 dataset. This suggests that AICL is not merely confined to adapting to a restricted and well-defined dataset but extends to effectively extracting and leveraging motion features to guide generation in more diverse and complex scenarios. The visual comparisons illustrated in Fig.\ref{Fig: Main-Result} further affirm the advantages of AICL. While the baseline model, VC, exhibits difficulties in accurately capturing and rendering the intended actions, and LVDM and VC2 present noticeable inconsistencies in frame sequences, the adoption of AICL facilitates all models in producing seamless motion and consistent frames. Notably, the distinct differences between the reference videos and the generated samples prove that AICL is not merely mimicking the reference content but learning and applying motion cues as intended, thereby enhancing the generative process.

\begin{figure}[h]
	\centering
	\includegraphics[width=\linewidth]{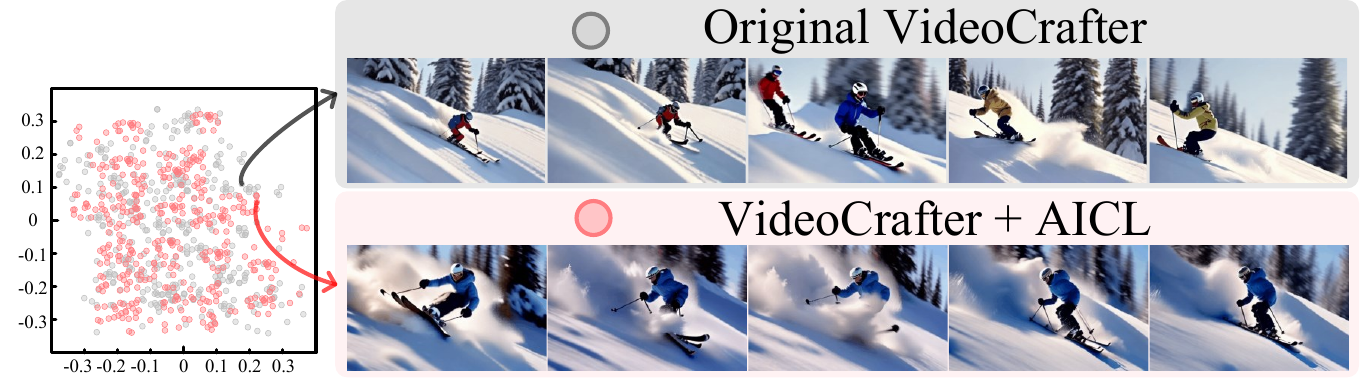}
	\caption{The visual of generated video feature by PCA.}
	\label{Fig: r-FVD}
\end{figure}

We have a marginal improvement on VideoCrafter because FVD calculates the distribution between the generated video and the dataset video. In Fig. ~\ref{Fig: r-FVD}, AICL improves action generation and maintains the original spatial layout, which leads to a minimal effect on the global distribution.

\begin{table}[h]
\caption{Performers on different speed action using LVDM + AICL (bigger number of speed group means faster action).}
\centering
\begin{tabular}{@{}c|ccccc@{}}
\toprule
Speed Group & \textcircled{1} & \textcircled{2} & \textcircled{3} & \textcircled{4} & \textcircled{5} \\ \midrule
FVD $\downarrow$ & 1057 & 984 & 938 & 1144 & 925 \\ \bottomrule
\end{tabular}%
\label{speed}
\end{table}

\noindent \textbf{Varying Degrees of Change.} To test AICL's performance on actions with varying degrees of change, we divide the actions in UCF-101 into 5-speed groups based on their magnitude of motion, and the results are shown in Table ~\ref{speed}. The magnitude of motion has a minor effect on performance, which demonstrates the robustness of AICL.

\noindent \textbf{Unseen Action.} Trained on open-domain datasets, it's hard to tell which action that current models have never seen at all. So we conduct novel action experiments by creating random word combinations (e.g., ``hssr'', ``tyuu'') as action names, and picking out some videos with the same actions as references. As the number of reference videos increases from 0 to 3, the FVD value decreases from 5476 to 3371 (improved 38.44\%), which demonstrates our method is good at generalizing to new actions.

\begin{figure}[h]
	\centering
	\includegraphics[width=\linewidth]{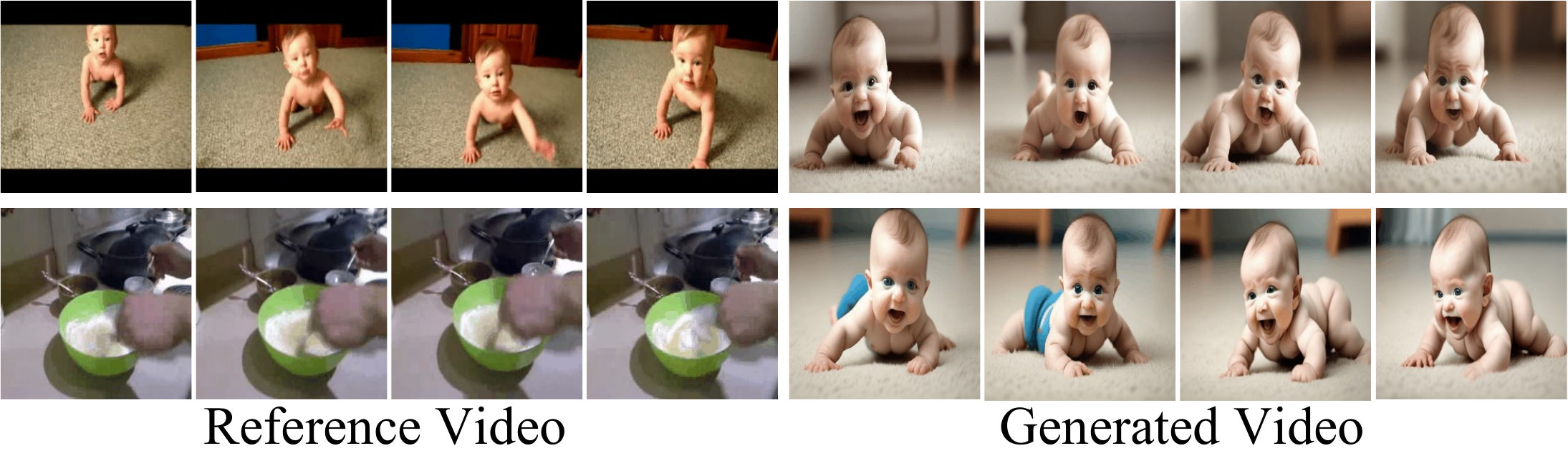}
	\caption{An example of wrong action reference video.}
	\label{Fig: r-same}
\end{figure}

\noindent \textbf{Open-Domain Problem.} For usability to End Users, our method is not limited to pre-defined actions proved above, and retrieval for reference videos is not a major issue for current techniques. As an example of ``baby crawl'' with the wrong reference video in Fig. \ref{Fig: r-same}, we demonstrate that even if the retrieval fails for a proper reference, the original performance will not be degraded.

\begin{figure}[h]
	\centering
	\includegraphics[width=\linewidth]{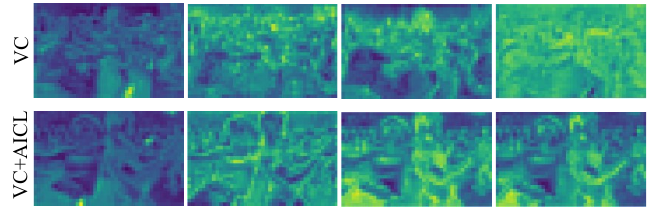}
	\caption{Visualization of attention score for cross attention in original video diffusion (up) and video diffusion model with AICL (down). The input text is ``Two people are dancing salsa''. Original video diffusion cannot be clearly understood action, and the attention area is incorrect. AICL can clearly understand the action.}
	\label{Fig: attention-score}
\end{figure}

\noindent \textbf{About Information Injected by Action Prism.} To better illustrate the role of our Action Prism module and its impact on generation, we visualized the attention in the Cross Attention mechanism of the original Diffusion video generation model and the Diffusion video generation model augmented with AICL. The attention is represented by the attention scores obtained from the product of Q and K in the attention mechanism. As shown in Fig. ~\ref{Fig: attention-score}, for actions that the original model fails to generate, the model without AICL cannot effectively extract information from the text for action generation. In contrast, the model with AICL can effectively model action information from the reference video.

\subsection{Generalization Study} \label{Sec: Generalization Study}

\begin{figure*}[h]
	\centering
	\includegraphics[width=\textwidth]{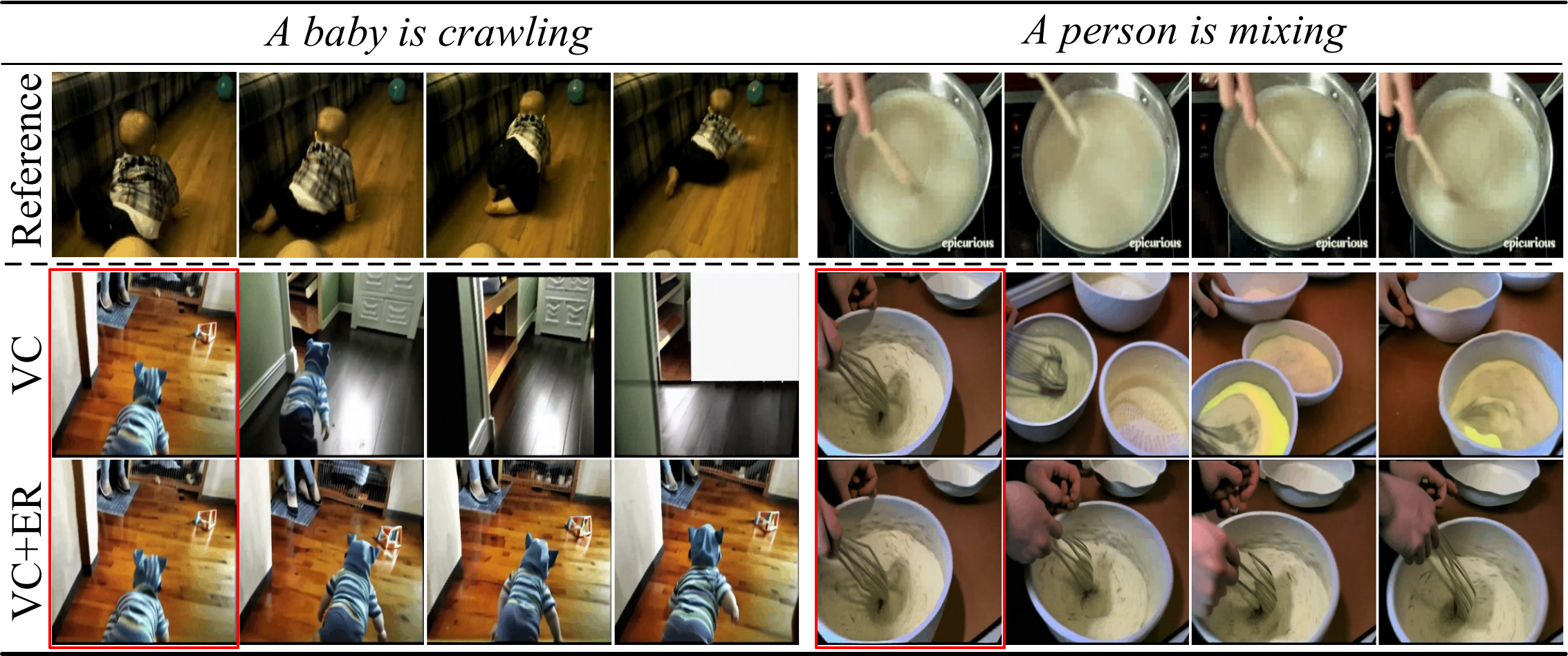}
	\caption{Qualitative results of AICL applied to VideoCrafter (VC) image-to-video using UCF-101 dataset. The condition image is notated by the red boxes.}
	\label{Fig: image}
\end{figure*}

\begin{figure*}[t]
	\centering
	\includegraphics[width=\textwidth]{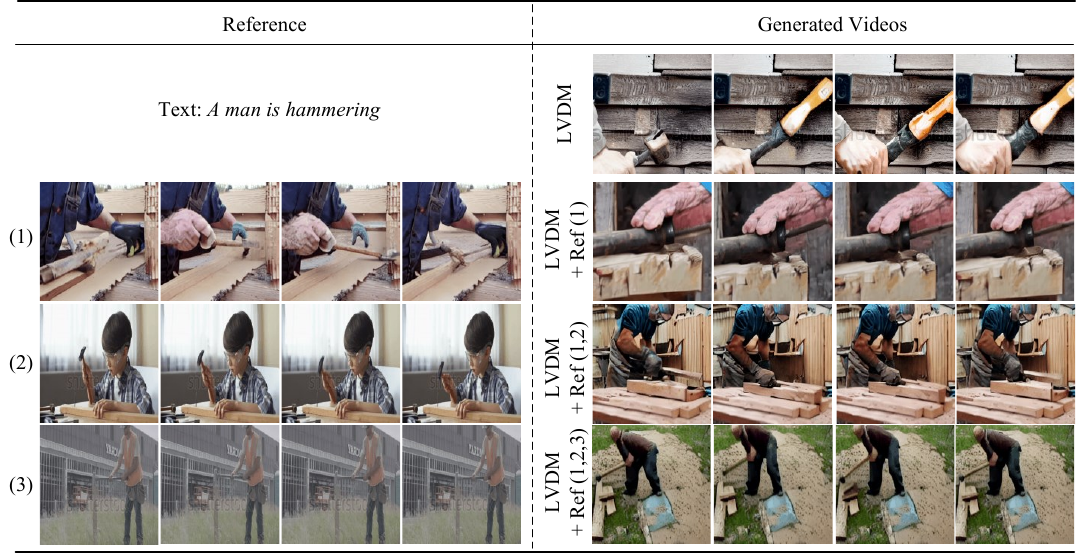}
	\caption{Qualitative results of different numbers of reference videos with AICL applied to LVDM using WebVid dataset.}
	\label{Fig: x-shot}
\end{figure*}

To assess the adaptability and broader applicability of AICL, we expand its functionality to encompass image-to-video synthesis. Specifically, AICL is integrated with VideoCrafter, forming a composite system capable of processing both image and textual inputs. This new model shares the same training method as our initial text-to-video framework. The efficacy of our method is evident from the quantitative assessments presented in Table \ref{Table: image}, as well as the qualitative evaluations depicted in Fig.\ref{Fig: image}. First, AICL significantly boosts the video generation performance across all metrics. Second, VideoCrafter generates inconsistent sequences and fails to depict motion accurately. When we attempt to use a classification network to recognize accuracy, the output accuracy with AICL is higher. AICL markedly enhances its capacity to produce coherent frames that are in harmony with the supplied image, and effectively captures the dynamic action ``crawling'' and ``mixing''.

\begin{table}[h]
    \caption{Generalization study of AICL applied to VideoCrafter image-to-video using UCF-101 dataset.}
    \centering{
            \begin{tabular}{@{}l|cccc@{}}
            \toprule
            Model      & CC $\uparrow$  & FC $\uparrow$  & FVD $\downarrow$ & ACC $\uparrow$\\ \midrule
            VC & 27.16          & 92.59          & 1132 & 64.32\%            \\
            VC + AICL  & \textbf{27.22} & \textbf{92.71} & \textbf{1125} & \textbf{82.94\%}    \\ \bottomrule
            \end{tabular}
    }
    \label{Table: image}
\end{table}

\begin{figure*}[t]
	\includegraphics[width=\textwidth]{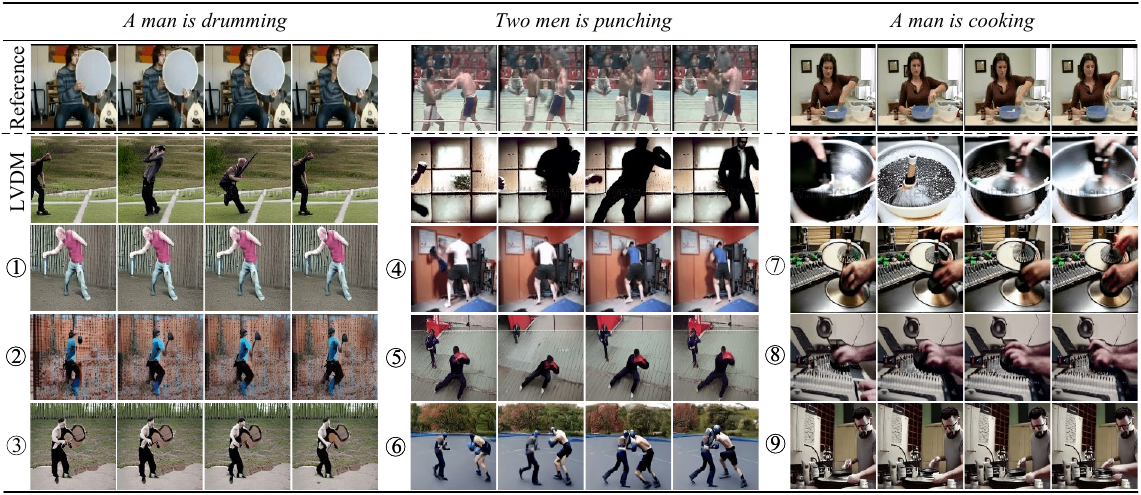}
	\caption{Qualitative results of ablation study of AICL applied to LVDM using UCF dataset. \textcircled{$x$} corresponds to the experiment model in Table ~\ref{Table: ablation}.}
	\label{Fig: sum-ablation}
\end{figure*}

\subsection{Ablation Study} \label{Sec: Ablation Study}

In this section, we study the specific design of our method, including the Action Prism, action integration, and number of reference videos. 

\noindent\textbf{Action Prism and Action Integration.} To ensure the best components of Action Prism and action integration, We build 7 baselines based on the LVDM, and to make the comparison clearer we notify them from \textcircled{1} to \textcircled{9}. For clarity in comparison, these baselines are enumerated from \textcircled{1} to \textcircled{9}, with \textcircled{3}, \textcircled{6}, and \textcircled{9} all representing our finalized AICL model. The quantitative results are demonstrated in Table \ref{Table: ablation} and the qualitative results are shown in Fig.\ref{Fig: sum-ablation}.

Initially, we analyze the integral components of the Action Prism. The baseline \textcircled{1}, which integrates solely cross-attention layers within the Action Prism, already marks a notable advancement over the vanilla LVDM. Augmenting this with self-attention layers, as seen in the model \textcircled{3}, further amplifies motion distillation through enhanced internal information extraction. A comparison between \textcircled{2} and \textcircled{3} reveals that altering the interplay between visual and conditional data within the cross-attention mechanism disrupts motion understanding, leading to suboptimal text-visual alignment and video semantic generation. As illustrated in Fig.\ref{Fig: sum-ablation}, only model \textcircled{3} adeptly interprets and distills effective motion features for "drumming" action generation. Moreover, we delve into the impact of spatial and temporal cross-attention within the Action Prism, discovering that an exclusive focus on spatial cross-attention yields superior outcomes in terms of text-visual alignment, video coherence, and visual fidelity. We think this is attributed to the challenge of sparse information arrangement posed by decoupled spatial and temporal processing in capturing motion relevant to the conditioned input. Referring to Fig.\ref{Fig: sum-ablation}, the inclusion of temporal cross-attention falters in action generation.

Subsequently, we explore various strategies for feature infusion into the VDMs. Note that for the combination of spatial and temporal cross-attention in the model \textcircled{8}, we duplicate the input video for the Action Prism thus obtaining two independent features for spatial and temporal attention calculation which we find performs better than utilizing the same feature. Through the comparative analysis of models \textcircled{7}, \textcircled{8}, and \textcircled{9}, we find that spatial feature injection inadequately facilitates action generation and the complete conditional input, distorting the intended "two people" condition into an erroneous "single person" depiction in Fig.\ref{Fig: sum-ablation}. Furthermore, the improvement from spatial cross-attention in visual quality is marginal. Combining all results, we configure our AICL model in the optimal setting in other experiments.

\begin{table}[h]
    \caption{Quantitative results of different numbers of reference videos with AICL applied to LVDM using WebVid dataset.}
    \centering{
            \begin{tabular}{@{}c|ccc@{}}
            \toprule
            Numbers & CC $\uparrow$  & FC $\uparrow$  & FVD $\downarrow$ \\ \midrule
            0  & 30.42          & 94.53          & 1349             \\
            1  & 30.41          & \textbf{97.46} & 998              \\
            2  & 30.51          & 97.38          & \textbf{992}     \\
            3  & \textbf{30.52} & 97.29          & 1034             \\ \bottomrule
            \end{tabular}
        }
    \label{Table: x-shot}
\end{table}

\noindent\textbf{Number of Reference Videos.} To explore the performance of AICL with different numbers of reference videos, we directly use the trained model to test multiple reference videos without fine-tuning. 
In Table \ref{Table: x-shot}, models with AICL significantly outperform the base LVDM no matter how many reference videos are chosen. Totally, extending the number of references leaves minor impacts on the objective metrics. Observing Fig.\ref{Fig: x-shot}, with reference videos extended from the local (hand) to the global (body), the model gradually comprehends the action ``hammering'' in a more complete way, and follows this way to generate. Since the visual quality is comparable, we chose to select a single reference video for each generation in all other experiments.

\section{Conclusion and Discussion}
In this paper, we introduced AICL, a novel framework to augment the capabilities of pretrained video diffusion models in generating complex actions. AICL leverages readily available video content as references to guide the diffusion process, employing in-context learning. We also propose the Action Prism to extract critical motion-related features from comprehensive videos. These features are then seamlessly integrated into the generation process through the newly added cross-attention, ensuring that the enhancement does not compromise the integrity of the original video diffusion models. Extensive experiments demonstrate the efficiency of AICL in motion extraction and action generation without simple content duplication. This research paves the way for further exploration of in-context learning in video generation.

\textbf{Limitation.} Current AICL demonstrates proficient capabilities in motion imitation, yet it encounters limitations in enhancing the generation of objects involved in actions that pre-existing video diffusion models struggle to synthesize. For instance, accurately depicting the action of ``playing the violin'' necessitates a vivid representation of both the violin itself and the intricate motions associated with playing it. While our approach effectively guides the generation of motion based on references, it falls short in the detailed generation of the violin. This paper acknowledges that the integration of spatial cross-attention with temporal ones does not rectify this shortcoming. Addressing this challenge will be a primary focus of our future research.


%
%
\bibliographystyle{splncs04}
\bibliography{egbib}

\begin{thebibliography}{10}
\providecommand{\url}[1]{\texttt{#1}}
\providecommand{\urlprefix}{URL }
\providecommand{\doi}[1]{https://doi.org/#1}

\bibitem{alayrac2022flamingo}
Alayrac, J.B., Donahue, J., Luc, P., Miech, A., Barr, I., Hasson, Y., Lenc, K., Mensch, A., Millican, K., Reynolds, M., et~al.: Flamingo: a visual language model for few-shot learning. NeurIPS  (2022)

\bibitem{WebVid}
Bain, M., Nagrani, A., Varol, G., Zisserman, A.: Frozen in time: {A} joint video and image encoder for end-to-end retrieval. In: 2021 {IEEE/CVF} International Conference on Computer Vision, {ICCV} 2021, Montreal, QC, Canada, October 10-17, 2021 (2021)

\bibitem{bar2022visual}
Bar, A., Gandelsman, Y., Darrell, T., Globerson, A., Efros, A.: Visual prompting via image inpainting. NeurIPS  (2022)

\bibitem{chen2023videocrafter1}
Chen, H., Xia, M., He, Y., Zhang, Y., Cun, X., Yang, S., Xing, J., Liu, Y., Chen, Q., Wang, X., Weng, C., Shan, Y.: Videocrafter1: Open diffusion models for high-quality video generation (2023)

\bibitem{chen2024videocrafter2}
Chen, H., Zhang, Y., Cun, X., Xia, M., Wang, X., Weng, C., Shan, Y.: Videocrafter2: Overcoming data limitations for high-quality video diffusion models (2024)

\bibitem{chen2022improving}
Chen, M., Du, J., Pasunuru, R., Mihaylov, T., Iyer, S., Stoyanov, V., Kozareva, Z.: Improving in-context few-shot learning via self-supervised training. arXiv preprint arXiv:2205.01703  (2022)

\bibitem{ding2022gpt}
Ding, B., Qin, C., Liu, L., Chia, Y.K., Joty, S., Li, B., Bing, L.: Is gpt-3 a good data annotator? arXiv preprint arXiv:2212.10450  (2022)

\bibitem{DingZHT22cogview2}
Ding, M., Zheng, W., Hong, W., Tang, J.: Cogview2: Faster and better text-to-image generation via hierarchical transformers. In: NeurIPS (2022)

\bibitem{gu2023pre}
Gu, Y., Dong, L., Wei, F., Huang, M.: Pre-training to learn in context. arXiv preprint arXiv:2305.09137  (2023)

\bibitem{LVDM}
He, Y., Yang, T., Zhang, Y., Shan, Y., Chen, Q.: Latent video diffusion models for high-fidelity video generation with arbitrary lengths. CoRR  (2022)

\bibitem{hertz2022prompt}
Hertz, A., Mokady, R., Tenenbaum, J., Aberman, K., Pritch, Y., Cohen-Or, D.: Prompt-to-prompt image editing with cross attention control. arXiv preprint arXiv:2208.01626  (2022)

\bibitem{HoCSWGGKPNFS22ImagenVideo}
Ho, J., Chan, W., Saharia, C., Whang, J., Gao, R., Gritsenko, A.A., Kingma, D.P., Poole, B., Norouzi, M., Fleet, D.J., Salimans, T.: Imagen video: High definition video generation with diffusion models. arXiv preprint arXiv:2210.02303  (2022)

\bibitem{ho2020ddpm}
Ho, J., Jain, A., Abbeel, P.: Denoising diffusion probabilistic models. Advances in neural information processing systems  \textbf{33},  6840--6851 (2020)

\bibitem{Conditional-Latent-Perturbation}
Ho, J., Saharia, C., Chan, W., Fleet, D.J., Norouzi, M., Salimans, T.: Cascaded diffusion models for high fidelity image generation. J. Mach. Learn. Res.  (2022)

\bibitem{Diffusion-Guidance}
Ho, J., Salimans, T.: Classifier-free diffusion guidance. In: NeurIPS. vol. abs/2207.12598 (2022)

\bibitem{HongDZLT23CogVideo}
Hong, W., Ding, M., Zheng, W., Liu, X., Tang, J.: Cogvideo: Large-scale pretraining for text-to-video generation via transformers. In: ICLR (2023)

\bibitem{huang2023language}
Huang, S., Dong, L., Wang, W., Hao, Y., Singhal, S., Ma, S., Lv, T., Cui, L., Mohammed, O.K., Liu, Q., et~al.: Language is not all you need: Aligning perception with language models. arXiv preprint arXiv:2302.14045  (2023)

\bibitem{isabelle2002proceedings}
Isabelle, P., Charniak, E., Lin, D.: Proceedings of the 40th annual meeting of the association for computational linguistics. In: Proceedings of the 40th Annual Meeting of the Association for Computational Linguistics (2002)

\bibitem{2023dalle3}
James, B., Gabriel, G., Li, J., Tim, B., Jianfeng, W., Linjie, L., Long, O., Juntang, Z., Joyce, L., Yufei, G., Wesam, M., Prafulla, D., Casey, C., Yunxin, J., Aditya, R.: Improving image generation with better captions. https://cdn. openai. com/papers/dall-e-3  (2023)

\bibitem{xFormers2022}
Lefaudeux, B., Massa, F., Liskovich, D., Xiong, W., Caggiano, V., Naren, S., Xu, M., Hu, J., Tintore, M., Zhang, S., Labatut, P., Haziza, D., Wehrstedt, L., Reizenstein, J., Sizov, G.: xformers: A modular and hackable transformer modelling library. \url{https://github.com/facebookresearch/xformers} (2022)

\bibitem{li2023finding}
Li, X., Qiu, X.: Finding supporting examples for in-context learning. arXiv preprint arXiv:2302.13539  (2023)

\bibitem{magister2022teaching}
Magister, L.C., Mallinson, J., Adamek, J., Malmi, E., Severyn, A.: Teaching small language models to reason. arXiv preprint arXiv:2212.08410  (2022)

\bibitem{min2021metaicl}
Min, S., Lewis, M., Zettlemoyer, L., Hajishirzi, H.: Metaicl: Learning to learn in context. arXiv preprint arXiv:2110.15943  (2021)

\bibitem{moens2021findings}
Moens, M.F., Huang, X.J., Specia, L., Yih, W.t.: Findings of the association for computational linguistics: Emnlp 2021. In: EMNLP (2021)

\bibitem{ohya2013automatic}
Ohya, H., Morishima, S.: Automatic mash up music video generation system by remixing existing video content. In: 2013 International Conference on Culture and Computing. pp. 157--158. IEEE (2013)

\bibitem{openai2023gpt4}
OpenAI: Gpt-4: Generative pre-trained transformer 4. \url{https://openai.com/gpt-4/} (2023)

\bibitem{Clip}
Radford, A., Kim, J.W., Hallacy, C., Ramesh, A., Goh, G., Agarwal, S., Sastry, G., Askell, A., Mishkin, P., Clark, J., Krueger, G., Sutskever, I.: Learning transferable visual models from natural language supervision. In: ICML (2021)

\bibitem{rasley2020deepspeed}
Rasley, J., Rajbhandari, S., Ruwase, O., He, Y.: Deepspeed: System optimizations enable training deep learning models with over 100 billion parameters. Proceedings of the 26th ACM SIGKDD International Conference on Knowledge Discovery \& Data Mining  (2020)

\bibitem{Stable-Diffusion}
Rombach, R., Blattmann, A., Lorenz, D., Esser, P., Ommer, B.: High-resolution image synthesis with latent diffusion models. In: CVPR (2022)

\bibitem{rubin2021learning}
Rubin, O., Herzig, J., Berant, J.: Learning to retrieve prompts for in-context learning. arXiv preprint arXiv:2112.08633  (2021)

\bibitem{imagen}
Saharia, C., Chan, W., Saxena, S., Li, L., Whang, J., Denton, E.L., Ghasemipour, K., Gontijo~Lopes, R., Karagol~Ayan, B., Salimans, T., et~al.: Photorealistic text-to-image diffusion models with deep language understanding. NeurIPS  (2022)

\bibitem{satapathy2023video}
Satapathy, S.K., Parmar, D.: Video generation by summarizing the generated transcript. In: 2023 3rd Asian Conference on Innovation in Technology (ASIANCON). pp.~1--5. IEEE (2023)

\bibitem{Diffusion}
Sohl{-}Dickstein, J., Weiss, E.A., Maheswaranathan, N., Ganguli, S.: Deep unsupervised learning using nonequilibrium thermodynamics. In: Bach, F.R., Blei, D.M. (eds.) ICML (2015)

\bibitem{song2020ddim}
Song, J., Meng, C., Ermon, S.: Denoising diffusion implicit models. arXiv preprint arXiv:2010.02502  (2020)

\bibitem{UCF}
Soomro, K., Zamir, A.R., Shah, M.: {UCF101:} {A} dataset of 101 human actions classes from videos in the wild. CoRR  (2012)

\bibitem{sun2023exploring}
Sun, Y., Chen, Q., Wang, J., Wang, J., Li, Z.: Exploring effective factors for improving visual in-context learning. arXiv preprint arXiv:2304.04748  (2023)

\bibitem{tsimpoukelli2021multimodal}
Tsimpoukelli, M., Menick, J.L., Cabi, S., Eslami, S., Vinyals, O., Hill, F.: Multimodal few-shot learning with frozen language models. NeurIPS  (2021)

\bibitem{FVD}
Unterthiner, T., van Steenkiste, S., Kurach, K., Marinier, R., Michalski, M., Gelly, S.: Towards accurate generative models of video: {A} new metric {\&} challenges. CoRR  (2018)

\bibitem{VaswaniSPUJGKP17}
Vaswani, A., Shazeer, N., Parmar, N., Uszkoreit, J., Jones, L., Gomez, A.N., Kaiser, L., Polosukhin, I.: Attention is all you need. In: NeurIPS. pp. 5998--6008 (2017)

\bibitem{wang2023modelscope}
Wang, J., Yuan, H., Chen, D., Zhang, Y., Wang, X., Zhang, S.: Modelscope text-to-video technical report. arXiv preprint arXiv:2308.06571  (2023)

\bibitem{VideoFactory}
Wang, W., Yang, H., Tuo, Z., He, H., Zhu, J., Fu, J., Liu, J.: Videofactory: Swap attention in spatiotemporal diffusions for text-to-video generation. arXiv preprint arXiv:2305.10874  (2023)

\bibitem{wang2023videocomposer}
Wang, X., Yuan, H., Zhang, S., Chen, D., Wang, J., Zhang, Y., Shen, Y., Zhao, D., Zhou, J.: Videocomposer: Compositional video synthesis with motion controllability. arXiv preprint arXiv:2306.02018  (2023)

\bibitem{wang2023images}
Wang, X., Wang, W., Cao, Y., Shen, C., Huang, T.: Images speak in images: A generalist painter for in-context visual learning. In: CVPR (2023)

\bibitem{wang2022super}
Wang, Y., Mishra, S., Alipoormolabashi, P., Kordi, Y., Mirzaei, A., Arunkumar, A., Ashok, A., Dhanasekaran, A.S., Naik, A., Stap, D., et~al.: Super-naturalinstructions: Generalization via declarative instructions on 1600+ nlp tasks. arXiv preprint arXiv:2204.07705  (2022)

\bibitem{wei2021finetuned}
Wei, J., Bosma, M., Zhao, V.Y., Guu, K., Yu, A.W., Lester, B., Du, N., Dai, A.M., Le, Q.V.: Finetuned language models are zero-shot learners. arXiv preprint arXiv:2109.01652  (2021)

\bibitem{wei2023symbol}
Wei, J., Hou, L., Lampinen, A., Chen, X., Huang, D., Tay, Y., Chen, X., Lu, Y., Zhou, D., Ma, T., et~al.: Symbol tuning improves in-context learning in language models. arXiv preprint arXiv:2305.08298  (2023)

\bibitem{wilde1894}
Wilde, O.: Phrases and Philosophies for the Use of the Young. Publisher (1894)

\bibitem{wu2023tune}
Wu, J.Z., Ge, Y., Wang, X., Lei, S.W., Gu, Y., Shi, Y., Hsu, W., Shan, Y., Qie, X., Shou, M.Z.: Tune-a-video: One-shot tuning of image diffusion models for text-to-video generation. In: Proceedings of the IEEE/CVF International Conference on Computer Vision. pp. 7623--7633 (2023)

\bibitem{WuGWLGHSQS22tuneavideo}
Wu, J.Z., Ge, Y., Wang, X., Lei, W., Gu, Y., Hsu, W., Shan, Y., Qie, X., Shou, M.Z.: Tune-a-video: One-shot tuning of image diffusion models for text-to-video generation. arXiv preprint arXiv:2212.11565  (2022)

\bibitem{xu2023k}
Xu, B., Wang, Q., Mao, Z., Lyu, Y., She, Q., Zhang, Y.: $ k $ nn prompting: Beyond-context learning with calibration-free nearest neighbor inference. arXiv preprint arXiv:2303.13824  (2023)

\bibitem{xu2023small}
Xu, C., Xu, Y., Wang, S., Liu, Y., Zhu, C., McAuley, J.: Small models are valuable plug-ins for large language models. arXiv preprint arXiv:2305.08848  (2023)

\bibitem{yang2023enabling}
Yang, J., Bors, A.G.: Enabling the encoder-empowered gan-based video generators for long video generation. In: 2023 IEEE International Conference on Image Processing (ICIP). pp. 1425--1429. IEEE (2023)

\bibitem{zhang2023makes}
Zhang, Y., Zhou, K., Liu, Z.: What makes good examples for visual in-context learning? arXiv preprint arXiv:2301.13670  (2023)

\bibitem{MagicVideo}
Zhou, D., Wang, W., Yan, H., Lv, W., Zhu, Y., Feng, J.: Magicvideo: Efficient video generation with latent diffusion models. CoRR  (2022)

\bibitem{zhu2023moviefactory}
Zhu, J., Yang, H., He, H., Wang, W., Tuo, Z., Cheng, W.H., Gao, L., Song, J., Fu, J.: Moviefactory: Automatic movie creation from text using large generative models for language and images. arXiv preprint arXiv:2306.07257  (2023)

\end{thebibliography}

\end{document}